\documentclass[10pt,twocolumn,letterpaper]{article}

\usepackage{cvpr}              

\usepackage{graphicx}
\usepackage{amsmath}
\usepackage{amssymb}
\usepackage{booktabs}
\usepackage{colortbl}
\usepackage{makecell}
\usepackage{multirow,multicol}
\usepackage{subcaption}
\usepackage{wrapfig,lipsum,booktabs}
\usepackage{bbm}
\usepackage{pifont}
\usepackage{lipsum}
\usepackage{tabu}
\usepackage{duckuments}

\def\eg{\emph{e.g.}}
\def\ie{\emph{i.e.}}

\newcolumntype{L}[1]{>{\raggedright\let\newline\\\arraybackslash\hspace{0pt}}m{#1}}
\newcolumntype{C}[1]{>{\centering\let\newline\\\arraybackslash\hspace{0pt}}m{#1}}
\newcolumntype{R}[1]{>{\raggedleft\let\newline\\\arraybackslash\hspace{0pt}}m{#1}}

\usepackage{amsmath,amsfonts,bm}

\def\eqref#1{equation~\ref{#1}}

\def\1{\bm{1}}

\def\mE{{\bm{E}}}

\def\mR{{\bm{R}}}

\DeclareMathAlphabet{\mathsfit}{\encodingdefault}{\sfdefault}{m}{sl}
\SetMathAlphabet{\mathsfit}{bold}{\encodingdefault}{\sfdefault}{bx}{n}

\definecolor{cvprblue}{rgb}{0.21,0.49,0.74}
\usepackage[pagebackref,breaklinks,colorlinks,allcolors=cvprblue]{hyperref}

\def\paperID{10} 
\def\confName{CVPR}
\def\confYear{2026}

\title{Tempered Self-Similarity Alignment for Physically Plausible Video Generation}

\author{Manjin Kim \hspace{18mm} Suha Kwak \hspace{18mm} Minsu Cho \vspace{2mm}\\
Pohang University of Science and Technology (POSTECH)\vspace{2mm}\\\small{
\url{https://cvlab.postech.ac.kr/project/TSA}}
}

\begin{document}
\maketitle
\begin{abstract}
Despite remarkable advances in video generative models, they still struggle to generate physically realistic videos, frequently exhibiting appearance drift, implausible motion, and temporal inconsistencies.
In this work, we address this limitation by transferring relational knowledge encoded in spatio-temporal self-similarity (STSS) from visual foundation models into video generative models.
STSS represents pairwise similarities among features across space and time, revealing the relational structure of how objects interact with other entities throughout a video, effectively capturing real-world dynamics, including object motion and semantic transformations.
To transfer this relational knowledge, we propose Tempered Self-similarity Alignment (TSA) loss, which transforms STSS into probabilistic correspondence distributions and trains the video generative model to align its correspondence distributions with those of the visual foundation model on dynamically changing regions.
Evaluated on VideoPhy and VideoPhy2 benchmarks, our method demonstrates substantial improvements in physical plausibility across diverse interaction scenarios, validating the effectiveness of transferring relational knowledge for physically realistic video generation.
\end{abstract}

\section{Introduction}
\label{sec:intro}
\begin{figure}[t]
    \centering
        \centering
        \includegraphics[width=\linewidth]{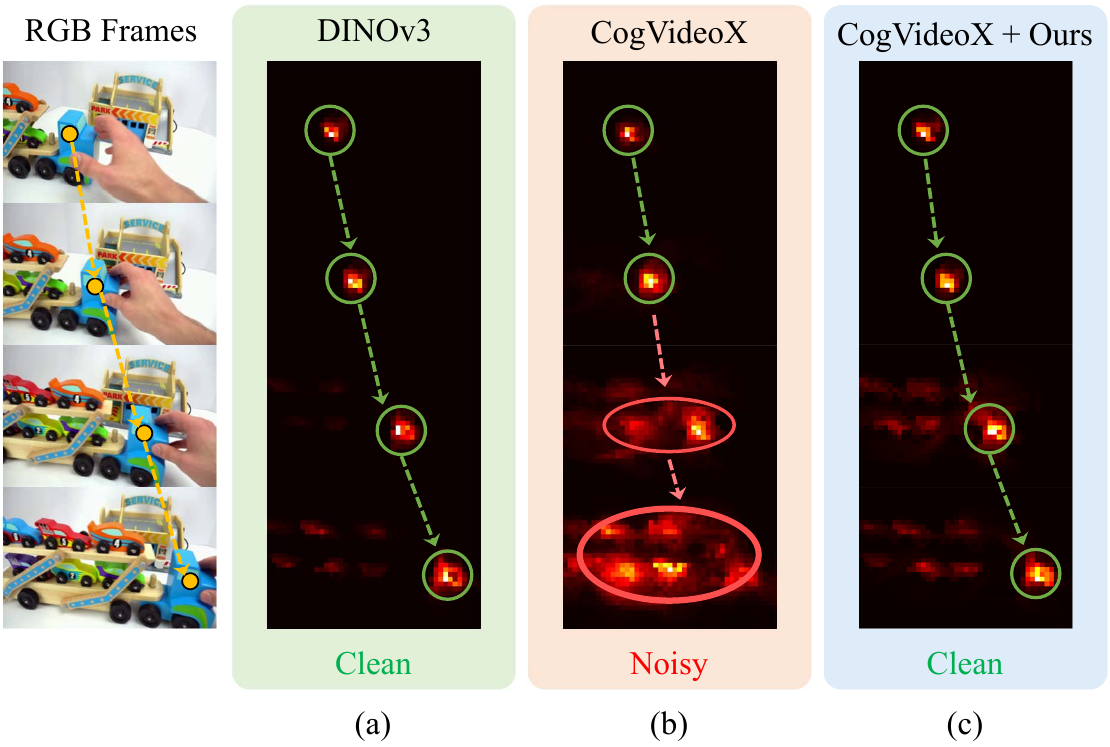}
    \vspace{-7mm}
    \caption{\textbf{Aligning spatio-temporal correspondence improves realistic dynamics in video generation.} Given the same video input, we compare spatio-temporal correspondence probability maps obtained from (a) a visual foundation model and (b) a video generative model. The foundation model captures clear correspondences, whereas the generative model produces noisy and inaccurate ones. (c) We here transfer the accurate spatio-temporal correspondences from the foundation model to the generative model, guiding the model toward more realistic motion dynamics.}
    \label{fig:teaser}
    \vspace{-4mm}
\end{figure}




Recent advances in video diffusion models~\cite{svd, hunyuanvideo, cogvideox, wan2025wan, pyramidalflow, brooks2024video, causvid} have enabled the generation of high-fidelity and high-resolution videos from text prompts. However, despite these remarkable achievements, current video generation models still struggle to produce physically plausible videos, frequently exhibiting appearance drift, unrealistic deformations, and inconsistent motion dynamics~\cite{videophy, videophy2, phygenbench, physicsiq}. These failures in physical realism limit the applicability of video generation models in domains requiring accurate motion simulation, such as robotics training, virtual environment construction, and video prediction.

Several approaches have been proposed to improve physical realism in video generation.
One line of work leverages physics simulators to synthesize videos with physically accurate dynamics~\cite{liu2024physgen, lin2024phy124, lin2024phys4dgen, zhang2024physdreamer, xie2024physgaussian, liu2025unleashing, montanaro2024motioncraft, xie2025physanimator}.
While effective, such simulation-based methods rely on heavy simulators and are typically restricted to specific physical domains, making them difficult to scale to diverse real-world scenarios.
Another direction utilizes multimodal large language models (MLLMs) to reason about physical properties and guide the generation process toward improved physical consistency~\cite{liu2025unleashing, lin2024phys4dgen, xue2025phyt2v, wang2025wisa}.
However, these approaches often depend on iterative refinement processes that significantly increase generation latency and their effectiveness is inherently bounded by the physical reasoning capability of the MLLMs.
A third line of work exploits auxiliary motion signals such as optical flow~\cite{track4gen, videojam, moalign, montanaro2024motioncraft}.
Although optical flow provides useful motion cues, it primarily captures short-term displacements between adjacent frames, making it difficult to provide supervision for long-range motion dynamics or structural changes over time.

In this work, we propose to leverage spatio-temporal self-similarity (STSS)~\cite{shechtman2005space, selfsimilarity, selfy} as a richer supervisory signal for physically plausible video generation.
STSS, \ie, pairwise similarities among visual features across spacetime, suppresses photometric variations such as color and texture while emphasizing relational structures among visual entities in videos.
Such relational patterns naturally encode how objects move, deform, and interact with surrounding entities across the entire video, capturing physically meaningful dynamics.

To effectively inject the relational knowledge encoded in STSS into video generation models, we introduce \textit{Tempered Self-similarity Alignment (TSA)} loss, which aligns STSS captured by pre-trained visual foundation models with that of video generative models.
However, directly aligning raw STSS tensors often yields overly smooth correspondence distributions, providing ambiguous supervision for precise motion learning.
Instead, we temper the STSS by applying temperature-scaled normalization to obtain sharpened correspondence probability distributions, which we then align across the two models.
These distributions represent object motion as \textit{a soft trajectory across the entire video}.
Unlike optical flow, which represents displacement as a one-hot correspondence restricted adjacent frames, this formulation captures long-range dynamics and inherent uncertainties in structural transformations.
Furthermore, we introduce background masking that restricts this alignment exclusively to dynamic regions, enabling the model to focus on physically grounded motion dynamics while avoiding unnecessary overhead on static regions.
We evaluate our method on VideoPhy~\cite{videophy} and VideoPhy2~\cite{videophy2} and demonstrate that TSA loss effectively improves the physical realism of generated videos.

Our main contributions are summarized as follows:
\begin{itemize}
    \item We propose Tempered Self-similarity Alignment (TSA) loss, which effectively transfers relational knowledge in STSS from visual foundation models into video diffusion models for enhancing realistic dynamics.
    \item We introduce a tempering mechanism that transforms STSS into sharpened correspondence probability distributions, enabling more precise motion supervision than direct STSS alignment.
    \item We demonstrate that our method significantly improves physical plausibility on VideoPhy and VideoPhy2 benchmarks, achieving state-of-the-art performance in generating physically realistic videos.
\end{itemize}


\section{Related Work}
\label{sec:related_work}

\noindent \textbf{Physically Plausible Video Generation.}
Despite recent advances in video diffusion models~\cite{svd, hunyuanvideo, cogvideox, wan2025wan, pyramidalflow, brooks2024video, causvid}, they still struggle to produce physically plausible videos, often exhibiting identity drift,  unrealistic deformations, or inconsistent dynamics~\cite{videophy, videophy2, phygenbench, physicsiq}.
To address these limitations, recent studies have explored various strategies to enhance physical realism in video generation.
One line of research integrates external physics simulators, such as MPM~\cite{stomakhin2013material, daviet2016semi, ram2015material}, with generative models to synthesize physically grounded videos~\cite{liu2024physgen, lin2024phy124, lin2024phys4dgen, zhang2024physdreamer, xie2024physgaussian, liu2025unleashing, montanaro2024motioncraft, xie2025physanimator}.
Another direction leverages LLMs for physical property analysis and reasoning about physical phenomena, guiding video generation toward improved physical consistency~\cite{liu2025unleashing, lin2024phys4dgen, xue2025phyt2v, wang2025wisa}.
Additionally, several methods utilize optical flow as conditioning or guidance signals to generate more realistic motion~\cite{videojam, track4gen, montanaro2024motioncraft}.
Recently, several methods leverage representation alignment~\cite{repa} to transfer physical knowledge from visual foundation models into video diffusion models~\cite{videorepa, crepa, geometryforcing}.
Building on this direction, we propose an alignment loss that transfers spatio-temporal self-similarities from visual foundation models into video generative models.

\noindent \textbf{Spatio-Temporal Self-Similarity.}
Self-similarity~\cite{shechtman2005space, selfsimilarity}, defined as pairwise similarities among visual features, effectively reveals underlying relational patterns in visual data, \eg, structural layouts or object motions, while suppressing appearance variations.
In the image domain, self-similarity has been employed as a relational descriptor for template matching~\cite{shechtman2005space,selfsimilarity}, capturing view-invariant geometric patterns~\cite{junejo2010view,junejo2008cross}, and establishing semantic correspondences~\cite{kang2021relational, kim2017fcss, torabi2013local}.
In videos, spatio-temporal self-similarity (STSS) has been leveraged to capture temporal dynamics such as motion.
Early works~\cite{junejo2010view,junejo2008cross} introduced temporal self-similarity descriptors for action recognition under viewpoint changes.
More recent methods compute spatial cross-similarities between adjacent frames to extract short-term motion cues~\cite{corrnet, motionsqueeze}, while others directly learn to transform STSS representations into bi-directional motion features~\cite{selfy, atm}.
Some methods~\cite{rsa,structvit} explicitly incorporate STSS into self-attention mechanisms for motion-centric video representation learning.
While these approaches primarily utilize STSS for video understanding such as action recognition, we extend its application to video generation.
Specifically, we propose to align STSS from visual foundation models with video diffusion models, transferring their knowledge of realistic motion dynamics to enable more physically plausible video generation.

\noindent \textbf{Knowledge Distillation \& Representation Alignment.}
Knowledge distillation (KD)~\cite{kd,zagoruyko2016paying} transfers knowledge from a teacher network to a student network by minimizing distance between their output representations. Beyond matching individual features, relation-based KD methods~\cite{rkd,spkd,rrd,vrm} extend KD to align relations among features, leading to more robust knowledge transfer.
Representation alignment (REPA)~\cite{repa} extends KD to generative models, aligning the noisy hidden states of diffusion models with the semantic representations of pre-trained visual encoders, effectively stabilizing and accelerating training.
This approach is initially proposed in image generation~\cite{repa, repae} and subsequently extended to video generation by several methods~\cite{videorepa, crepa, geometryforcing}.
Among them, VideoREPA~\cite{videorepa} proposes to distill STSS from video foundation models to improve physical realism in generated videos.
However, it directly aligns raw STSS tensors uniformly across all regions, limiting precise motion supervision.
Aligned with the spirit of relation-based KD, we re-interpret STSS as motion and introduce a tempered normalization to transform STSS into sharpened correspondence distributions. By aligning these distributions, our method provides more precise motion supervision than raw STSS alignment.
Furthermore, we restrict the alignment to dynamic regions, enabling more focused and effective motion learning compared to uniform spatial alignment.


\section{Preliminaries}

\begin{figure*}[t]
    \centering
    \includegraphics[width=\linewidth]{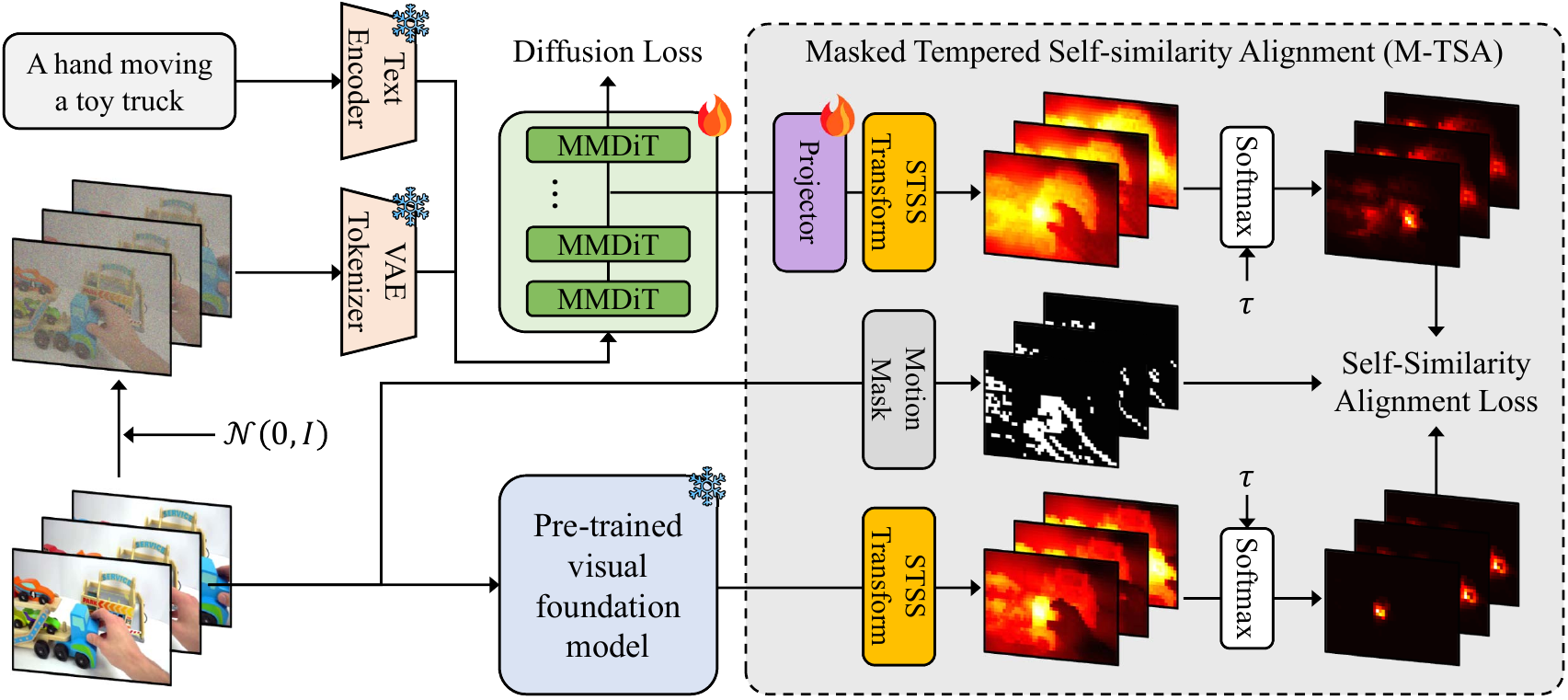}
    \caption{\textbf{Framework overview.} Our method aligns the noisy spatio-temporal correspondences of a video diffusion model with the accurate correspondences of a visual foundation model, guiding the generative model toward more realistic motion dynamics. By restricting this alignment to dynamic regions, it encourages motion-focused alignment for physically plausible video generation.}
    \label{fig:main}
    \vspace{-2mm}
\end{figure*}

\subsection{Video Diffusion Models}

Recent text-to-video diffusion models~\cite{svd, hunyuanvideo, cogvideox, wan2025wan, pyramidalflow} follow the Latent Diffusion Model (LDM) framework~\cite{ldm}, where the generation process is performed in the latent space of a pretrained Variational Autoencoder (VAE).
Given a video $\mathbf{V} \in \mathbb{R}^{F' \times H' \times W' \times 3}$, the VAE encoder compresses it into a latent representation $\mathbf{Z}^{(0)} \in \mathbb{R}^{F \times H \times W \times C}$ with reduced spatial and temporal resolutions.
A forward diffusion process corrupts $\mathbf{Z}^{(0)}$ by gradually injecting Gaussian noise over timesteps $t \in \{1, \ldots, T\}$, producing noisy latents $\mathbf{Z}^{(t)} = \alpha_t \mathbf{Z}^{(0)} + \sigma_t \epsilon$, where $\epsilon \sim \mathcal{N}(0, I)$ and $\alpha_t, \sigma_t$ are schedule-dependent coefficients.
A denoising network, typically a spatio-temporal transformer such as MM-DiT~\cite{stablediffusion3} in CogVideoX~\cite{cogvideox}, is trained to reverse this process by estimating the noise added at each timestep.
The training objective minimizes the mean squared error between the true and predicted noise as,
\begin{equation}
    \mathcal{L}_{\text{diffusion}} = \mathbb{E}_{t, \mathbf{Z}^{(0)}, \epsilon} \left[ \left\| \epsilon - \epsilon_\theta(\mathbf{Z}^{(t)}, t, c) \right\|_2^2 \right],
\end{equation}
where $c$ denotes text prompt.
At inference, the model starts from sampling pure noise $\mathbf{Z}^{(T)} \sim \mathcal{N}(0, I)$ and iteratively refines it through the learned reverse process to obtain $\hat{\mathbf{Z}}^{(0)}$. The VAE decoder then transforms $\hat{\mathbf{Z}}^{(0)}$ into the final video.
Applied to this LDM-based video diffusion model, we aim to enhance the physical plausibility of generated videos.

\subsection{Representation Alignment}
\label{subsec:repa}

While diffusion-based generative models produce high-quality samples, their internal representations tend to be less expressive and discriminative compared to those of large-scale visual foundation models trained (\eg, DINOv2~\cite{dinov2}).
To bridge this representational gap, representation alignment (REPA)~\cite{repa} has been proposed as a regularization method that transfers rich semantic knowledge from the pre-trained foundation models to the diffusion models.

REPA is initially introduced for image diffusion models~\cite{repa, repae}. Let the denoising network $f_\theta$ take a noisy image $\mathbf{Z}^{(t)}$ as input, conditioned on context $\mathbf{c}$ and timestep $t$. The network produces intermediate activations at its $l$-th layer $\mathbf{H} = f_\theta(\mathbf{Z}^{(t)}, t, \mathbf{c}, l) \in \mathbb{R}^{H \times W \times C_{\text{d}}}$, where $C_{\text{d}}$ is the channel dimension of the denosing network.
A pretrained visual encoder $g_\phi$ maps a clean image $\mathbf{X}$ to visual features $\mathbf{E} = g_\phi(\mathbf{X}) \in \mathbb{R}^{H \times W \times C_{\text{e}}}$, with feature dimension $C_{\text{e}}$.
To align the representations of the diffusion model with those of the visual foundation model, REPA introduces a lightweight projector $h_\psi$ that maps $\mathbf{H}$ to the encoder's feature space. The alignment objective encourages feature-wise consistency as:
\begin{equation}
\mathcal{L}_{\text{REPA}} = -\mathbb{E}_{\mathbf{X}, \mathbf{Z}^{(t)}, t} \left[ \frac{1}{HW} \sum_{i=1,j=1}^{H,W} \text{sim}\left(\mathbf{E}_{i,j}, h_\psi(\mathbf{H}_{i,j})\right) \right],
\end{equation}
where $\text{sim}(\cdot, \cdot)$ computes cosine similarity between corresponding patches. This regularizer is combined with the standard diffusion loss:
\begin{equation}
\mathcal{L}_{\text{total}} = \mathcal{L}_{\text{diffusion}} + \lambda \mathcal{L}_{\text{REPA}},
\end{equation}
where $\lambda$ balances generation fidelity and feature alignment.
Empirically, this approach accelerates training convergence and enhances perceptual quality by leveraging semantic priors from the pretrained encoder.
While REPA aligns feature representations directly, our method aligns probabilistic distributions of spatio-temporal correspondences derived from STSS, enabling more effective transfer of physically grounded motion dynamics.

\section{Methods}
\label{sec:methods}

Spatio-temporal self-similarity (STSS) effectively captures temporal dynamics such as motion by describing the relational structure across space and time in videos.
By transferring such STSS structures from visual foundation models into video diffusion models, we can guide the generative process toward producing physically realistic dynamics.
To this end, we propose \textit{Tempered Self-similarity Alignment (TSA)}, which converts the STSS tensor of each query token into probabilistic distributions of spatio-temporal correspondences of the query across frames, and aligns these correspondence distributions from visual foundation models with those from the video diffusion model.
We further restrict this alignment to dynamic regions, focusing the supervision on motion-salient areas where physically meaningful dynamics occur.

\subsection{Tempered Self-Similarity Alignment}

Given an input video $\mathbf{V} \in \mathbb{R}^{F' \times H' \times W' \times 3}$, we first extract spatio-temporal features $\mathbf{E}^{\text{VFM}}$ using a pretrained visual foundation model $g_\phi$.
We then obtain intermediate features from the $l$-th layer of the denoising network $f_\theta$ and pass them through a lightweight projector $h_\psi$ that maps the diffusion features into the pretrained encoder's feature space with matched spatio-temporal resolution, yielding $\mathbf{E}^{\text{VDM}}$:
\begin{align}
    \mathbf{E}^{\text{VFM}} &= g_\phi(\mathbf{V}) \in \mathbb{R}^{F \times H \times W \times C} \\
    \mathbf{E}^{\text{VDM}} &= h_\psi(f_\theta(\mathbf{Z}^{(t)}, t, c, l)) \in \mathbb{R}^{F \times H \times W \times C}.
\end{align}
We apply L2 normalization along the channel dimension and flatten the spatio-temporal dimensions to obtain $\bar{\mE}^{\text{VFM}}, \bar{\mE}^{\text{VDM}} \in \mathbb{R}^{N \times C}$, where $N = F H W$.
We then compute pairwise cosine similarity matrices that encode STSS relations:
\begin{align}
\mR^{\text{VFM}} = \bar{\mE}^{\text{VFM}} (\bar{\mE}^{\text{VFM}})^\top \in \mathbb{R}^{N \times N}, \\
\mR^{\text{VDM}} = \bar{\mE}^{\text{VDM}} (\bar{\mE}^{\text{VDM}})^\top \in \mathbb{R}^{N \times N},
\end{align}
where each row $\mR_i$ encodes the similarities between token $i$ and all other tokens across space and time.

To obtain frame-wise correspondences, we reshape $\mR \in \mathbb{R}^{N \times N}$ to $\mathbb{R}^{N \times F \times HW}$ and apply a temperature-scaled softmax along the spatial dimension:
\begin{equation}
\mathbf{P}^{\text{VFM}}_{i,f} = \text{softmax}\left(\frac{\mR^{\text{VFM}}_{i,f}}{\tau}\right), \mathbf{P}^{\text{VDM}}_{i,f} = \text{softmax}\left(\frac{\mR^{\text{VDM}}_{i,f}}{\tau}\right),
\end{equation}
where $\mathbf{P}_{i,f} \in \mathbb{R}^{HW}$ represents the probability distribution over spatial positions in the $f$-th frame that query token $i$ corresponds to.
The TSA loss aligns these correspondence distributions between the visual foundation model and the video diffusion model by minimizing the KL divergence:
\begin{equation}
\mathcal{L}_{\text{TSA}} = \frac{1}{NF} \sum_{i=1}^{N} \sum_{f=1}^{F} D_{\mathrm{KL}}\left(\mathbf{P}^{\text{VFM}}_{i,f} \,\|\, \mathbf{P}^{\text{VDM}}_{i,f}\right).
\end{equation}

\begin{figure}[t]
    \centering
    \scalebox{0.91}{
    \includegraphics[width=\linewidth]{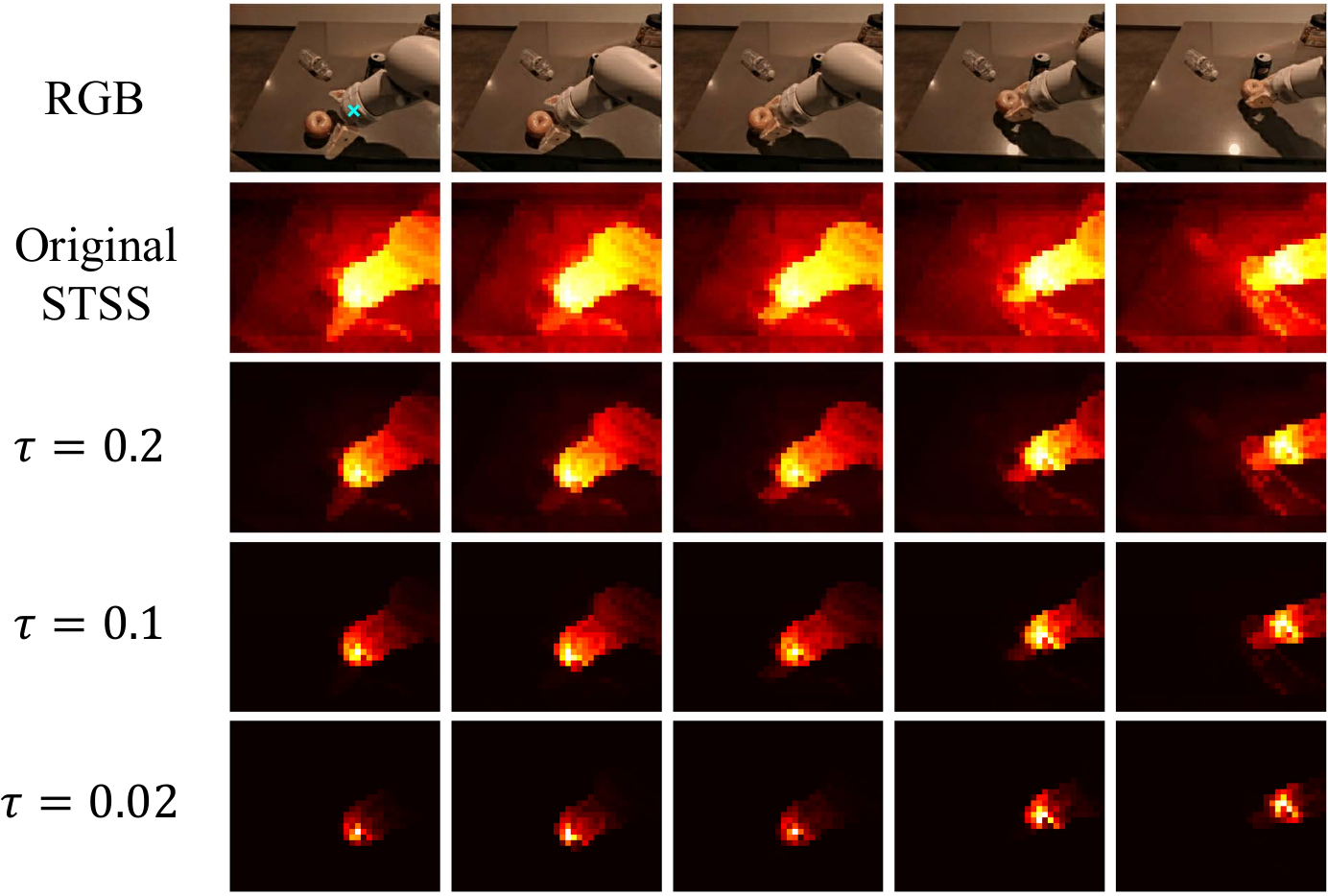}
    }
    \vspace{-2mm}
    \caption{\textbf{Controlling correspondence granularity by temperature.} As temperature gets lower, correspondence changes from object-level to part-level to point-level.}
    \label{fig:temperature}
    \vspace{-3mm}
\end{figure}

TSA offers two key advantages. First, it provides stronger signals for misaligned regions while attenuating gradients for well-aligned correspondences. The gradient of the TSA loss with respect to $\mR^{\text{VDM}}_{i,f}$ is computed as,
\begin{equation}
\frac{\partial \mathcal{L}_{\text{TSA}}}{\partial \mR^{\text{VDM}}_{i,f}} = \frac{1}{\tau}\left(\mathbf{P}^{\text{VDM}}_{i,f} - \mathbf{P}^{\text{VFM}}_{i,f}\right),
\end{equation}
which is proportional to the probability mismatch between the two distributions, unlike L1-based objectives~\cite{videorepa} that apply uniform gradients regardless of alignment quality.
Second, the temperature $\tau$ enables fine-grained motion supervision through correspondence sharpening.
Without temperature scaling, directly aligning raw STSS tensors~\cite{videorepa} yields overly smooth distributions that fail to provide precise motion supervision for each query token (Fig.~\ref{fig:temperature}, 2nd row).
By lowering $\tau$ to sharpen the correspondence distributions, TSA provides more precise spatial supervision per query, facilitating fine-grained motion learning while preserving object part-level structural information encoded in STSS (Fig.~\ref{fig:temperature}, 4th row).

\subsection{Motion-focused Self-Similarity Alignment}

While TSA effectively aligns STSS across the entire video, most regions in videos are static and lack meaningful temporal dynamics. Aligning correspondences in such static areas introduces unnecessary computational overhead and may hinder the model from focusing on dynamic regions where physically grounded motion cues exist. To address this inefficiency and improve motion modeling, we introduce \textit{Masked TSA (M-TSA)}, which reinforces STSS alignment on such dynamic regions.

Given an input video $\mathbf{V} \in \mathbb{R}^{F' \times H' \times W' \times 3}$, we first divide the video into non-overlapping tubulets of size $(P_\text{t}, P_\text{h}, P_\text{w})$, resulting in patch tokens $\mathbf{Y} \in \mathbb{R}^{F \times H \times W \times 3P_\text{t} P_\text{h} P_\text{w}}$. We compute the temporal difference across frames and calculate L1 norm for each patch token as,
\begin{equation}
\Delta_{t,h,w} = 
\begin{cases}
    \left\| \mathbf{Y}_{t,h,w} - \mathbf{Y}_{t-1,h,w} \right\|_1, & t \in \{2, \ldots, F\} \\
    \Delta_{2,h,w}, & t = 1
\end{cases}
\end{equation}
We identify dynamic regions by selecting the top-$k$ (e.g., $k=20\%$) tokens with the largest temporal differences for each frame, constructing a motion-saliency mask $\mathcal{M}$:
\begin{align}
\mathcal{M} &= \big\{I(t,h,w)\ \big|\ \Delta_{t,h,w}\ge r_t, \forall t \in \{1, \cdots, F\} \big\},
\end{align}

\noindent where $I(\cdot)$ denotes the index of the patch token in the flattened dimension ($N = F \times H \times W$), and $r_t = \text{top-}k\big(\{\Delta_{t,h,w}\}_{h=1,w=1}^{H,W}\big)$ denotes the threshold that retains the highest $k\%$ of temporal differences at the $t$-th frame. The M-TSA loss then performs STSS alignment exclusively to these motion-salient tokens:
\begin{equation}
\mathcal{L}_{\text{M-TSA}} = \frac{1}{|\mathcal{M}|F} \sum_{i\in\mathcal{M}} \sum_{f=1}^{F} D_{\mathrm{KL}}\left(\mathbf{P}^{\text{VFM}}_{i,f} \,\|\, \mathbf{P}^{\text{VDM}}_{i,f}\right).
\end{equation}
This selective alignment reduces computational and memory overhead by excluding static regions and enables the model to focus on capturing physically meaningful motions.
We combine the M-TSA loss with the diffusion loss for fine-tuning video diffusion models as,
\begin{equation}
    \mathcal{L}_{\text{total}} = \mathcal{L}_{\text{diffusion}} + \lambda \mathcal{L}_{\text{M-TSA}}.
\end{equation}

\begin{figure}[t]
    \centering
    \includegraphics[width=\linewidth]{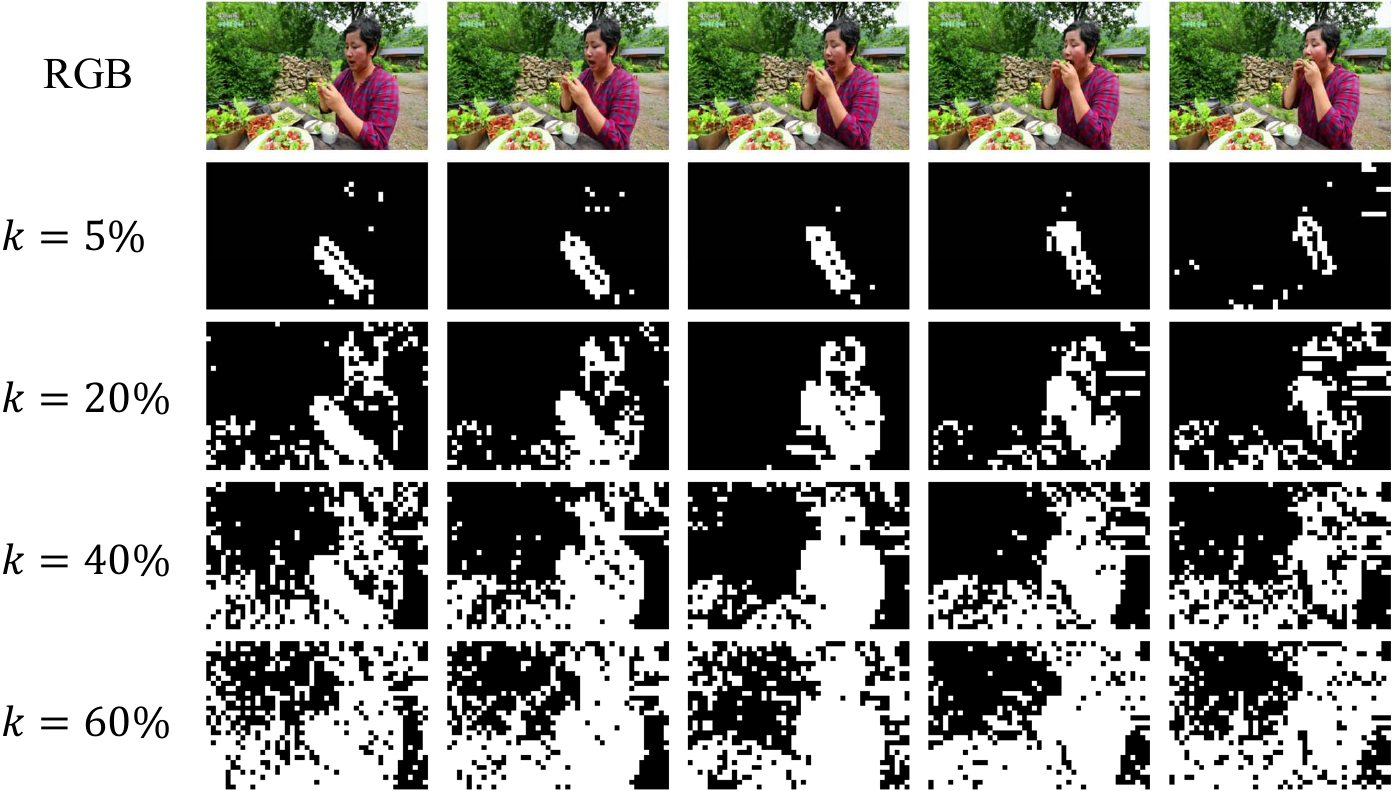}
    \vspace{-6mm}
    \caption{\textbf{Motion-saliency masks at different thresholds $k$.} Increasing $k$ highlights a larger proportion of the most dynamic regions in the video.}
    \label{fig:motion_mask}
    \vspace{-3mm}
\end{figure}


\section{Experiments}
\label{sec:experiments}
\subsection{Implementation Details}
\label{sec:implementation_details}


\begin{table*}[t]
    \centering
    \caption{Quantitative results on VideoPhy benchmark. We report Semantic Adherence (SA) and Physical Commonsense (PC) scores as the percentage of videos achieving a score $\geq 0.5$ across different interaction categories. $*$ finetuned on OpenVid subset.}
    \vspace{-2mm}
    \label{tab:videophy}
    \begin{tabular}{l|cc|cc|cc|cc}
        \Xhline{0.8pt}
    & \multicolumn{2}{c|}{Overall} & \multicolumn{2}{c|}{Solid-Solid} & \multicolumn{2}{c|}{Solid-Fluid} & \multicolumn{2}{c}{Fluid-Fluid} \\
    Method & SA & PC & SA & PC & SA & PC & SA & PC \\
    \Xhline{0.3pt}
    Cosmos-Diffusion-7B~\cite{agarwal2025cosmos} & 57.0 & 18.0 & - & - & - & - & - & -  \\
    DreamMachine~\cite{dreammachine} & 57.5 & 21.8 & 55.1 & 21.7 & 59.6 & 23.3 & 58.2 & 18.2  \\
    VideoCrafter2~\cite{videocrafter2} & 50.3 & 29.7 & 50.4 & 32.2 & 50.7 & 27.4 & 48.1 & 29.1  \\
    HunyuanVideo~\cite{hunyuanvideo} & 60.2 & 28.2 & 55.2 & 16.1 & 67.1 & 30.1 & 54.5 & 54.5  \\
    CogVideoX-2B~\cite{cogvideox} & 59.9 & 25.0 & 44.8 & 15.4 & 71.9 & 22.6 & 67.3 & 56.4 \\
    CogVideoX-2B + PhyT2V~\cite{xue2025phyt2v} & 42 & 29 & - & - & - & - & - & -  \\
    \Xhline{0.3pt}
    CogVideoX-2B$^*$ & 59.6 & 25.3 & 45.5 & 15.4 & 71.2 & 23.3 & 65.5 & 56.4 \\
    +$\mathcal{L}_{\text{REPA}}$~\cite{repa} & 60.5 & 22.7 & 45.5 & 9.1 & 72.6 & \textbf{27.4} & 67.3 & 45.5 \\
    +$\mathcal{L}_{\text{TRD}}$~\cite{videorepa} & 63.1 & 27.6 & 49.7 & 18.2 & \textbf{74.0} & \textbf{27.4} & 69.1 & 52.7 \\
    \rowcolor{gray!20}
    +$\mathcal{L}_{\text{TSA}}$ (ours) & 62.8 & 29.1 & 51.8 & 19.6 & 70.6 & 26.0 & \textbf{70.9} & 61.8 \\
    \rowcolor{gray!20}
    +$\mathcal{L}_{\text{M-TSA}}$ (ours) & \textbf{64.5} & \textbf{30.8} & \textbf{53.9} & \textbf{21.0} & \textbf{74.0} & \textbf{27.4} & 67.3 & \textbf{65.5} \\
    \Xhline{0.8pt}
    \end{tabular}
\end{table*}


\begin{table}[t]
    \centering
    \caption{Quantitative results on VideoPhy2. We report SA, PC, and Joint scores as the percentage of videos achieving scores $\geq 4$.}
    \vspace{-2mm}
    \label{tab:videophy2}
    \begin{tabular}{l|ccc}
    \Xhline{0.8pt}
    Method & Joint & SA & PC \\
    \Xhline{0.3pt}
    CogVideoX-2B & 22.4 & 27.1 & 65.9 \\
    CogVideoX-2B$^*$ & 22.9 & 26.8 & 68.0 \\
    +$\mathcal{L}_{\text{REPA}}$~\cite{repa} & 23.3 & 27.5 & 66.8  \\
    +$\mathcal{L}_{\text{TRD}}$~\cite{videorepa} & 23.1 & 27.0 & 68.4 \\
    \rowcolor{gray!20}
    +$\mathcal{L}_{\text{TSA}}$ (ours) & 24.0 & \textbf{28.2} & 69.3 \\
    \rowcolor{gray!20}
    +$\mathcal{L}_{\text{M-TSA}}$ (ours) & \textbf{24.4} & \textbf{28.2} & \textbf{69.8} \\
    \Xhline{0.8pt}
    \end{tabular}
\end{table}

We adopt CogVideoX-2B~\cite{cogvideox} and VideoMAEv2-B~\cite{videomaev2} as the base text-to-video diffusion model and visual foundation model, respectively.
We employ a lightweight projector composed of a 3-layer MLP followed by a 3D convolutional layer, which adjusts the spatio-temporal resolution of the diffusion model's representations to match that of the foundation model.
We set the loss weight $\lambda$, temperature $\tau$, and masking ratio $k$ to 0.5, 0.1, and 20, respectively.
We subsample 64K videos from OpenVid~\cite{openvid} for finetuning~\cite{videorepa}.
We fully finetune the model using AdamW~\cite{adamw} with a learning rate of $2 \times 10^{-6}$ and batch size of 32 for 4000 iterations on eight NVIDIA RTX 6000 Ada GPUs.

\subsection{Benchmarks}
\label{sec:datasets}
\noindent\textbf{VideoPhy}~\cite{videophy} is a benchmark designed to evaluate the physical plausibility of video generation models.
It consists of 344 prompts involving interactions between various material types in the physical world, categorized into solid-solid, solid-fluid, and fluid-fluid interactions.
The benchmark measures two metrics: \textit{Semantic Adherence (SA)}, which assesses whether the generated video faithfully follows the text prompt, and \textit{Physical Commonsense (PC)}, which evaluates whether the generated video exhibits physically realistic dynamics.
We use the VideoConPhysics auto-evaluator to compute scores in the range $[0, 1]$ and report the proportion of samples with scores $\geq 0.5$.
For evaluation, we use the upsampled captions provided by VideoREPA~\cite{videorepa}.

\noindent\textbf{VideoPhy2}~\cite{videophy2} extends the evaluation to human-object interactions, in contrast to the material-centric focus of VideoPhy.
It comprises 590 text prompts and measures SA and PC on an integer scale from 1 to 5.
Additionally, the Joint score is reported as the proportion of samples where both SA $\geq 4$ and PC $\geq 4$.
For automated evaluation, we employ the VideoPhy2-AutoEval model and utilize the upsampled prompts officially released by the benchmark~\cite{videophy2}.

\noindent\textbf{VBench}~\cite{vbench} is a comprehensive benchmark suite for evaluating video generative models. It consists of 946 text prompts and evaluates two aspects: video quality, which measures the perceptual quality of the video independent to the text prompt, and video–condition consistency, which evaluates how faithfully the video aligns with the text prompt. Each aspect comprises 8 metrics, resulting in a total of 16 metrics. In addition, we report the aggregated Quality Score, Semantic Score, and the Total Score to summarize performance across these aspects. For evaluation, we use the upsampled prompts provided by NOVA~\cite{nova}.

\subsection{Results}
\label{sec:results}

\noindent\textbf{VideoPhy.}
We evaluate our method on VideoPhy and present the results in Tab.~\ref{tab:videophy}.
Compared to the baseline CogVideoX-2B, applying TSA significantly improves the PC score, achieving 29.1\% overall, which surpasses VideoREPA (27.6\%).
This improvement demonstrates the effectiveness of our tempered self-similarity alignment approach, which provides fine-grained motion supervision and stronger gradients for misaligned regions through KL divergence minimization.
Adding background masking (M-TSA) further boosts performance, reaching an overall PC score of 30.8\%, outperforming PhyT2V~\cite{xue2025phyt2v} (29.0\%) that iteratively refines videos using physical knowledge from LLMs.
The largest improvement is observed in the Solid-Solid category, where PC increases from 15.4\% to 21.0\% (36.4\% relative improvement), with consistent gains also observed in Solid-Fluid (17.6\%) and Fluid-Fluid (16.1\%) interactions.
These results validate that focusing self-similarity alignment on dynamic regions enables more effective learning of physically grounded motion dynamics.

\noindent\textbf{VideoPhy2.}
We evaluate our method on VideoPhy2, a benchmark focused on human-object interactions, and summarize the results in Tab.~\ref{tab:videophy2}.
Consistent with the VideoPhy results, TSA outperforms VideoREPA, improving the Joint score from 23.1\% to 24.0\%.
Motion masking further enhances performance, with M-TSA achieving a Joint score of 24.4\%, demonstrating the consistent effectiveness of our approach across diverse human-object interaction scenarios.


\begin{table*}[t]
    \centering
    \caption{\textbf{Quantitative results on VBench}. $*$ finetuned on OpenVid subset.}
    \vspace{-2mm}
    \label{tab:vbench}
    \begin{subtable}{\linewidth}
        \setlength{\tabcolsep}{2pt}
        \centering
        \caption{Total scores and video quality metrics.}
        \label{tab:vbench:a}
        \scalebox{0.90}{
        \begin{tabular}{lcccccccccc}
        \Xhline{0.3pt}
        Method & \makecell{Total\\Score} & \makecell{Quality\\Score} & \makecell{Subject\\Consist.} & \makecell{Background\\Consist.} & \makecell{Temporal\\Flickering} & \makecell{Motion\\Smoothness} & \makecell{Dynamic\\Degree} & \makecell{Aesthetic\\Quality} & \makecell{Imaging\\Quality} & \makecell{Object\\Class} \\
        \Xhline{0.3pt}
        CogVideoX-2B* & 81.0 & 81.4 & 93.4 & 94.2 & \textbf{98.0} & 98.3 & \textbf{56.9} & \textbf{63.4} & 63.2 & \textbf{91.3} \\
        \rowcolor{gray!20} +$\mathcal{L}_{\text{M-TSA}}$ (ours) & \textbf{81.2} & \textbf{81.5} & \textbf{93.5} & \textbf{94.5} & \textbf{98.0} & \textbf{98.4} & 56.1 & 63.0 & \textbf{63.5} & 90.2 \\
        \Xhline{0.3pt}
        \end{tabular}
        }
    \end{subtable}

    \vspace{3mm}

    \begin{subtable}{\linewidth}
        \centering
        \caption{Video-condition consistency metrics.}
        \label{tab:vbench:b}
        \setlength{\tabcolsep}{3.45pt}
        \scalebox{0.90}{
        \begin{tabular}{lccccccccc}
        \Xhline{0.3pt}
        Method & \makecell{Semantic\\Score} & \makecell{Multiple\\Objects} & \makecell{Human\\Action} & Color & \makecell{Spatial\\Relationship} & Scene & \makecell{Appearance\\Style} & \makecell{Temporal\\Style} & \makecell{Overall\\Consistency} \\
        \Xhline{0.3pt}
        CogVideoX-2B* & 79.4 & 71.6 & 98.2 & \textbf{85.0} & 75.5 & 52.2 & 24.4 & \textbf{25.2} & 27.1 \\
        \rowcolor{gray!20} +$\mathcal{L}_{\text{M-TSA}}$ (ours) & \textbf{79.7} & \textbf{73.0} & \textbf{98.8} & 84.4 & \textbf{75.9} & \textbf{53.4} & \textbf{24.5} & \textbf{25.2} & \textbf{27.2} \\
        \Xhline{0.3pt}
        \end{tabular}
        }
    \end{subtable}
\end{table*}

\begin{figure*}[t]
    \centering
    \begin{subfigure}{0.325\linewidth}
        \centering
        \scalebox{1.0}{
        \includegraphics[width=\linewidth]{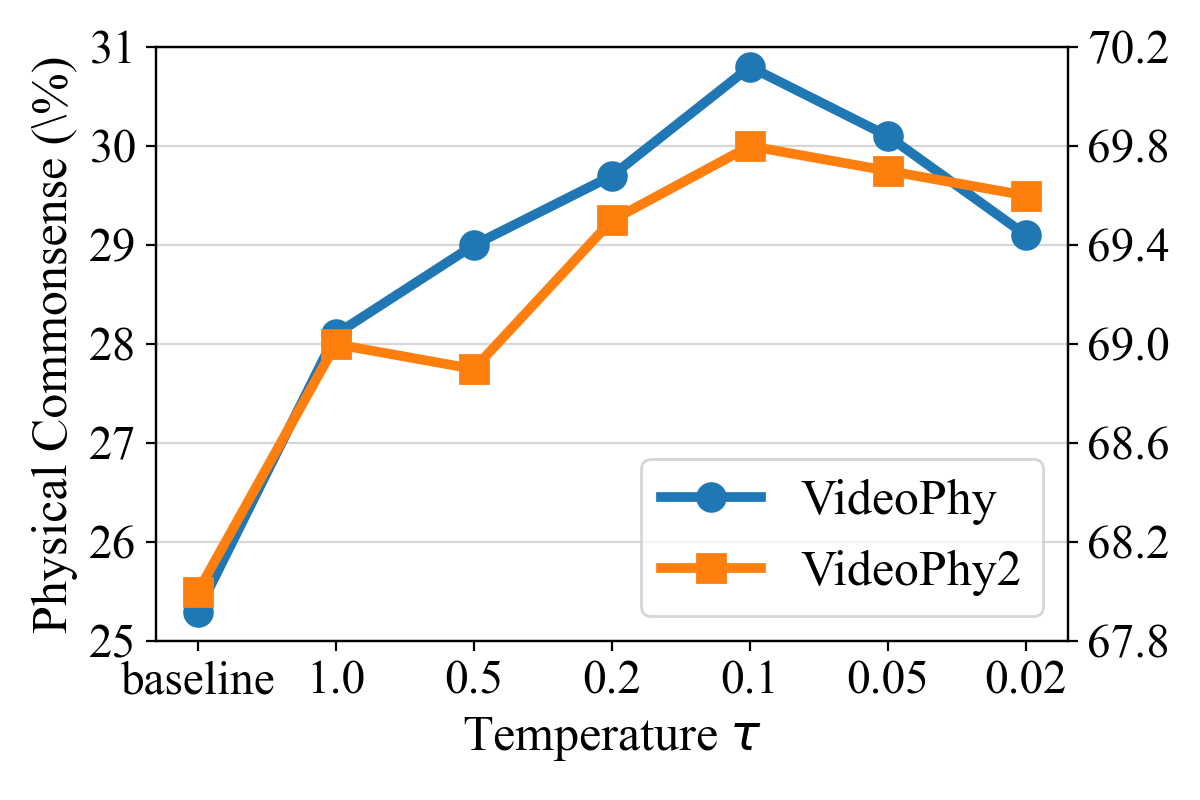}
        }
        \vspace{-5mm}
        \subcaption{Temperature $\tau$.}
        \label{fig:ablation_temperature}
    \end{subfigure}
    \begin{subfigure}{0.325\linewidth}
        \centering
        \scalebox{1.0}{
        \includegraphics[width=\linewidth]{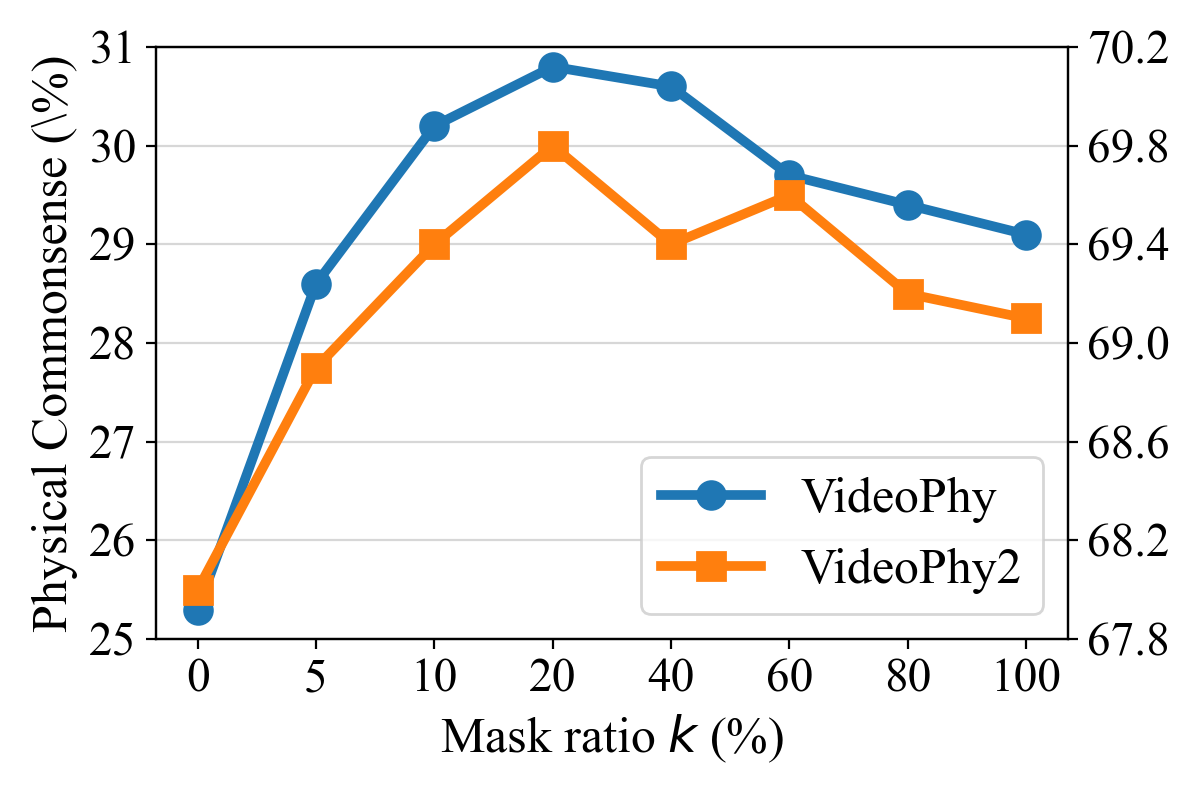}
        }
        \vspace{-5mm}
        \subcaption{Masking ratio $k$.}
        \label{fig:ablation_masking}
    \end{subfigure}
    \begin{subfigure}{0.325\linewidth}
        \centering
        \scalebox{1.0}{
        \includegraphics[width=\linewidth]{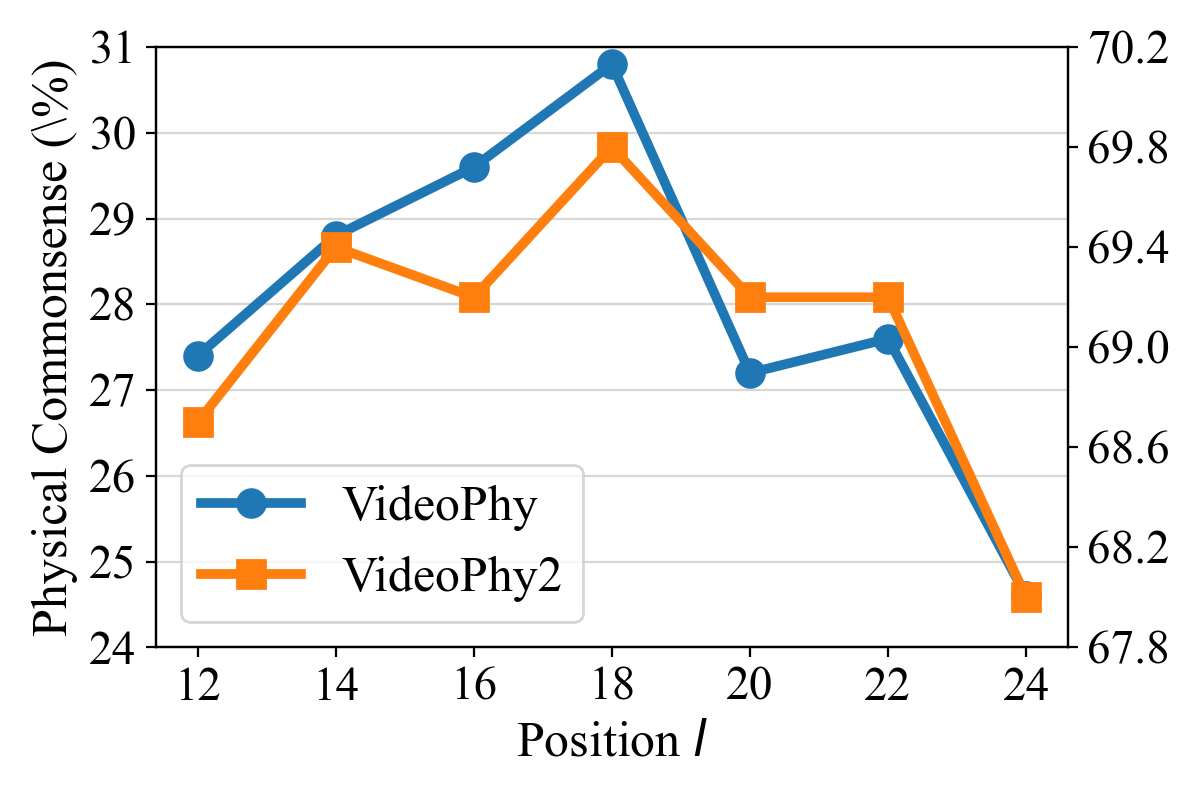}
        }
        \vspace{-5mm}
        \subcaption{Position $l$.}
        \label{fig:ablation_position}
    \end{subfigure}
    \vspace{-1mm}
    \caption{Ablation studies on VideoPhy. The PC scores are reported.}
    \label{fig:ablation}
\end{figure*}

\noindent\textbf{VBench.}
To ensure that our method preserves overall video quality while improving physical plausibility, we evaluate on VBench and present the results in Tab.~\ref{tab:vbench}.
Compared to the baseline, applying M-TSA preserves the overall video quality while yielding a slight improvement in the Total Score (81.0\%$\rightarrow$81.2\%).
This demonstrates that our method enhances physical plausibility without compromising the inherent video generation quality of the base model.

\subsection{Ablation Studies}
\label{sec:ablation_studies}

\noindent\textbf{Temperature.}
In Fig.~\ref{fig:ablation_temperature}, we evaluate the effect of temperature by gradually decreasing $\tau$ from 1.0 to lower values.
We observe clear performance improvements on both VideoPhy and VideoPhy2 when decreasing $\tau$ from 1.0 to 0.1, indicating that a sharper correspondence distribution encourages the model to capture more fine-grained temporal relations, which facilitates precise motion modeling.
However, lowering the temperature beyond this point leads to performance degradation.
We conjecture that excessively sharp distributions suppress broader structural cues—such as object part-level geometry—that are necessary for coherent motion understanding.
These results suggest that an appropriate balance between fine-grained motion cues and structural context is critical for effective self-similarity alignment.

\noindent\textbf{Background Masking.}
In Fig.~\ref{fig:ablation_masking}, we analyze the effect of the masking ratio $k$, which determines the proportion of dynamic regions selected for alignment. On VideoPhy, utilizing only the top $5\%$ of dynamic tokens yields performance superior to the baseline, with the optimal performance achieved at $k=20\%$. As the ratio increases beyond this point, we observe a gradual decline in performance. Consistent trends are observed on VideoPhy2. These results validate our hypothesis that aligning static regions introduces redundancy and computational noise. By restricting the alignment to motion-salient regions, our method effectively prevents the model from being biased toward static backgrounds, allowing it to concentrate its learning capacity on generating physically grounded motion dynamics.

\noindent\textbf{Alignment Position.}
In Fig.~\ref{fig:ablation_position}, we evaluate the effectiveness of feature alignment across various depths of the denoising network. We identify the 18-th transformer block of CogVideoX as the optimal layer for injecting the correspondence priors on both benchmarks and set as the default for our main experiments.

\noindent\textbf{Different Architectures.}
To demonstrate the architectural generality of our approach, we apply our loss to NOVA~\cite{nova}, an auto-regressive video generation model.
In Tab.~\ref{tab:nova}, our method consistently improves PC scores across all material types, despite structural differences between bi-directional diffusion and causal auto-regressive models.
This validates that our method is model-agnostic, broadly applicable to a wide range of video generative frameworks.


\begin{table}[t]
    \centering
    \setlength\tabcolsep{2.5pt}
    \caption{\textbf{Effect with NOVA~\cite{nova} on VideoPhy}. $*$ finetuned on OpenVid subset.}
    \vspace{-2mm}
    \label{tab:nova}
    \scalebox{0.90}{
    \begin{tabular}{lcccccccc}
    \Xhline{0.3pt}
    \multirow{2}{*}{method} & \multicolumn{2}{c}{overall} & \multicolumn{2}{c}{solid-solid} & \multicolumn{2}{c}{solid-fluid} & \multicolumn{2}{c}{fluid-fluid} \\
    & SA & PC & SA & PC & SA & PC & SA & PC \\
    \Xhline{0.3pt}
    NOVA-0.6B~\cite{nova}       & 42.7 & 21.2 & 27.3 & 12.6 & 54.8 & 23.3 & 50.9 & 38.2 \\
    NOVA-0.6B$^*$   & 44.5 & 20.1 & 31.5 & 12.6 & 55.5 & 21.2 & 49.1 & 36.4 \\
    +$\mathcal{L}_{\text{TRD}}$~\cite{videorepa} & \textbf{45.9} & 22.7 & 30.8 & 12.6 & \textbf{56.2} & 23.3 & \textbf{58.2} & 47.3 \\
    \rowcolor{gray!20}
    +$\mathcal{L}_{\text{TSA}}$ (ours) & 45.1 & 29.4 & 31.5 & 18.9 & 53.4 & 30.1 & \textbf{58.2} & 54.5 \\
    \rowcolor{gray!20}
    +$\mathcal{L}_{\text{M-TSA}}$ (ours) & 45.1 & \textbf{30.5} & \textbf{32.2} & \textbf{19.6} & 53.4 & \textbf{30.8} & 56.4 & \textbf{58.2} \\
    \Xhline{0.3pt}
    \end{tabular}
    }
\end{table}

\begin{figure*}[t]
    \centering
    \begin{subfigure}{\linewidth}
        \centering
        \includegraphics[width=\linewidth]{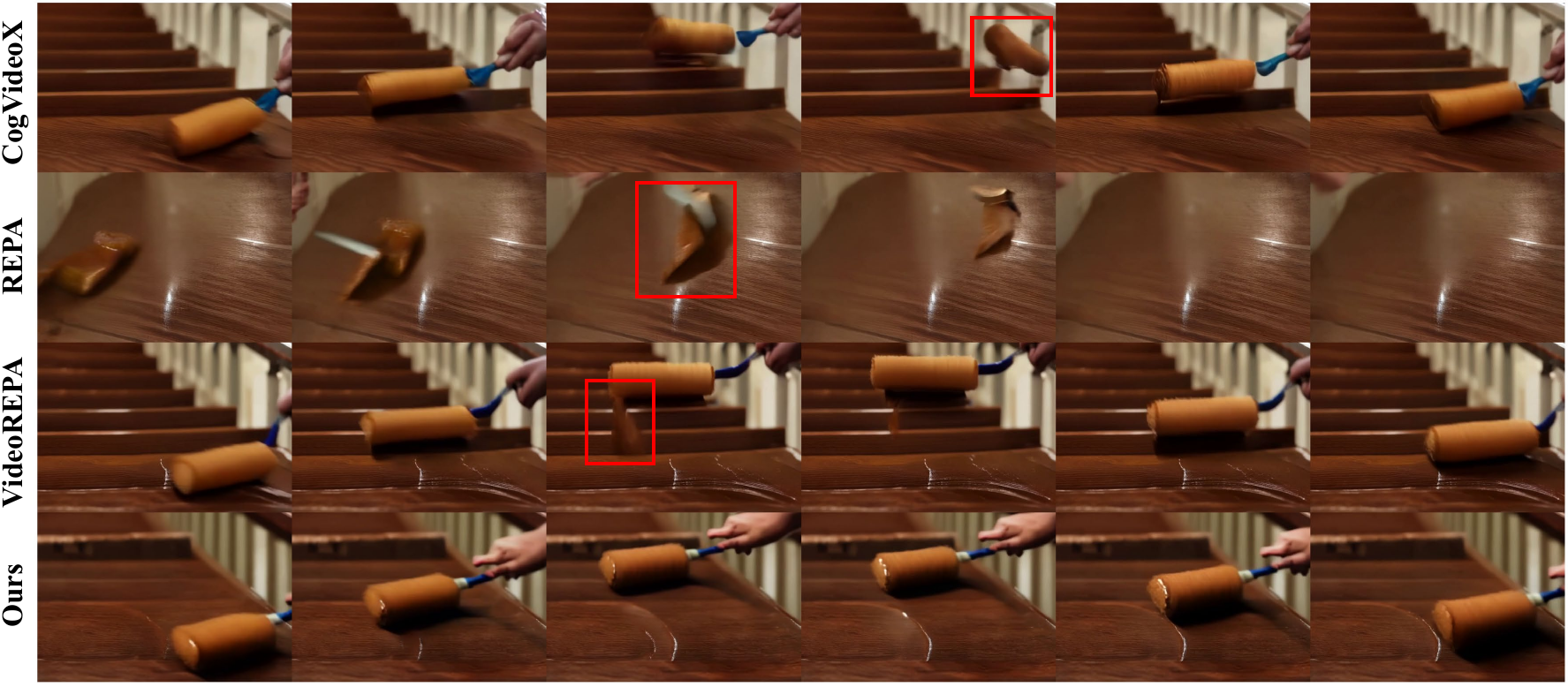}
        \subcaption{Prompt: A paint roller is used to apply a coat of brown paint ...}
        \label{fig:qual_1}
    \end{subfigure}

    \vspace{3mm}

    \begin{subfigure}{\linewidth}
        \centering
        \includegraphics[width=\linewidth]{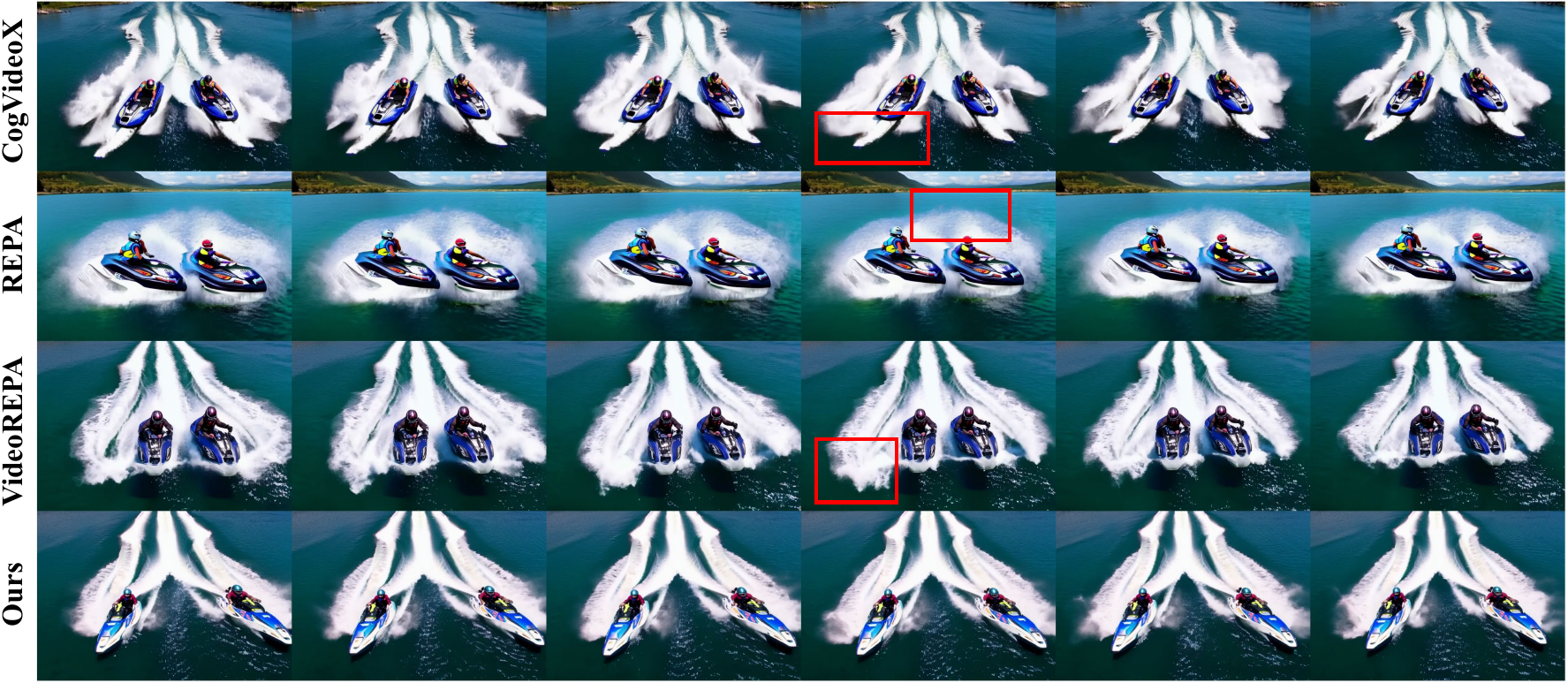}
        \subcaption{Prompt: Two jetskis race side-by-side, their wakes colliding ...}
        \label{fig:qual_2}
    \end{subfigure}


    \caption{\textbf{Qualitative results.} Red rectangles highlight regions with physically implausible or temporally inconsistent motion. Our method generates videos with physically realistic dynamics.}
    \label{fig:qual}
\end{figure*}

\subsection{Qualitative Results}
\label{sec:qualitative_results}
In Fig.~\ref{fig:qual}, we provide qualitative comparisons of videos generated by CogVideoX~\cite{cogvideox}, REPA~\cite{repa}, VideoREPA~\cite{videorepa}, and our method.
In Fig.~\ref{fig:qual_1}, existing methods exhibit appearance drift of the paint roller or generate unrealistic rolling motion, whereas our method preserves the object's appearance throughout the video while depicting realistic motion dynamics.
In Fig.~\ref{fig:qual_2}, other methods generate physically implausible water splashes that violate causality, \eg, splashes appearing in front of the jetski before it passes, while our method produces temporally coherent and physically plausible splash patterns.
These results validate that our approach effectively transfers realistic motion dynamics embedded in STSS to video generative models.
Please refer to Appendix~\ref{supp:additional_qual} for more qualitative results.


\section{Conclusion}
\label{sec:conclusion}

We have introduced Tempered Self-similarity Alignment (TSA), a novel framework that transfers spatio-temporal correspondences in visual foundation models into video diffusion models for physically plausible video generation.
By converting spatio-temporal self-similarity into probabilistic correspondence distributions and aligning such distributions exclusively on dynamically changing regions, our method enables effective alignment of physically grounded dynamics.
Through extensive ablation studies, we found that controlling correspondence granularity plays a critical role in motion learning and that filtering out static regions via motion masking substantially improves performance.
Our experiments on VideoPhy and VideoPhy2 demonstrate that TSA significantly improves the physical realism of generated videos in diverse real-world scenarios.

\noindent \textbf{Acknowledgements.}
This work was supported by Qualcomm-POSTECH UR (POH-621245) and Institute of Information \& Communications Technology Planning \& Evaluation (IITP) grants (RS-2022-II220959: Few-Shot Learning of Causal Inference in Vision and Language for Decision Making (40\%); RS-2022-II220264: Comprehensive Video Understanding and Generation with Knowledge-based Deep Logic Neural Network (30\%); RS-2024-00457882: National AI Research Lab Project (30\%)) funded by the Korea government (MSIT). 

{
    \small
    \bibliographystyle{ieeenat_fullname}
    \bibliography{main}
}

\clearpage
\appendix

\section{Additional Qualitative Results}
\label{supp:additional_qual}
\begin{figure*}[t]
    \centering
    \begin{subfigure}{\linewidth}
        \centering
        \includegraphics[width=\linewidth]{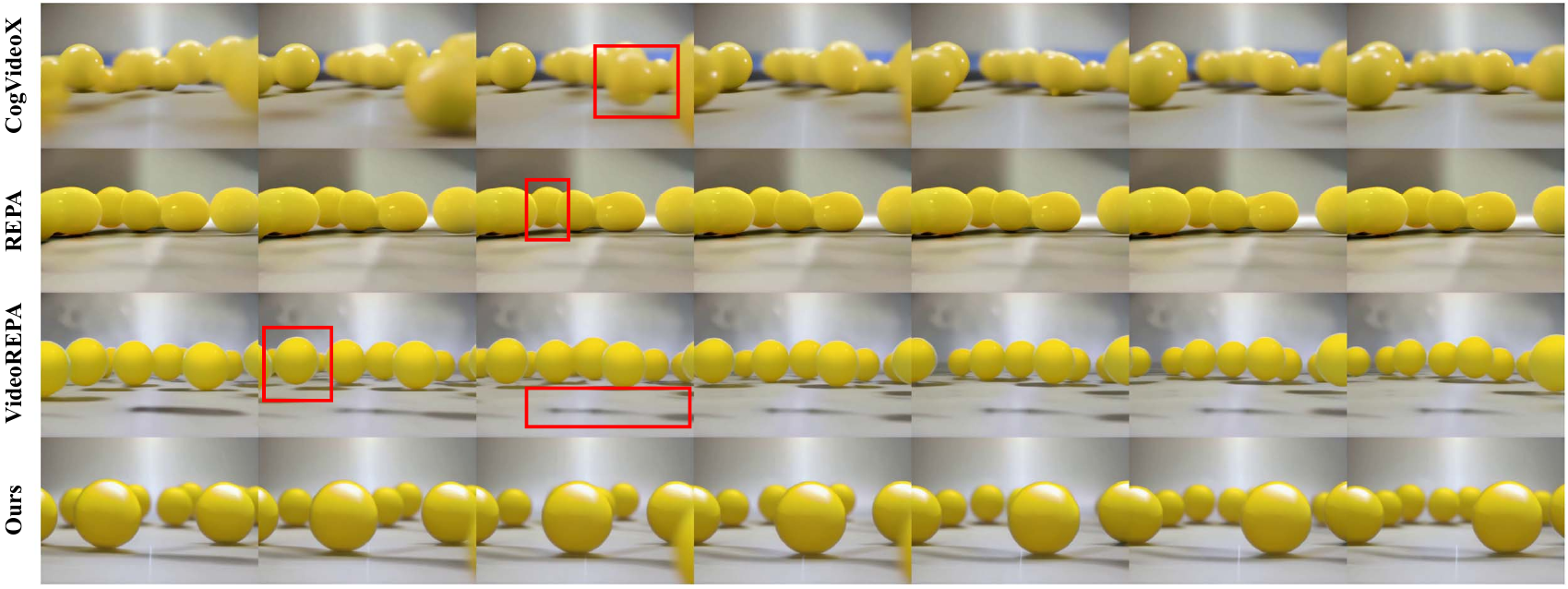}
        \subcaption{Prompt: Multiple ping pong balls are shown, illustrating different stages of a rally.}
        \label{fig:supp_qual_1}
    \end{subfigure}

    \vspace{2mm}

    \begin{subfigure}{\linewidth}
        \centering
        \includegraphics[width=\linewidth]{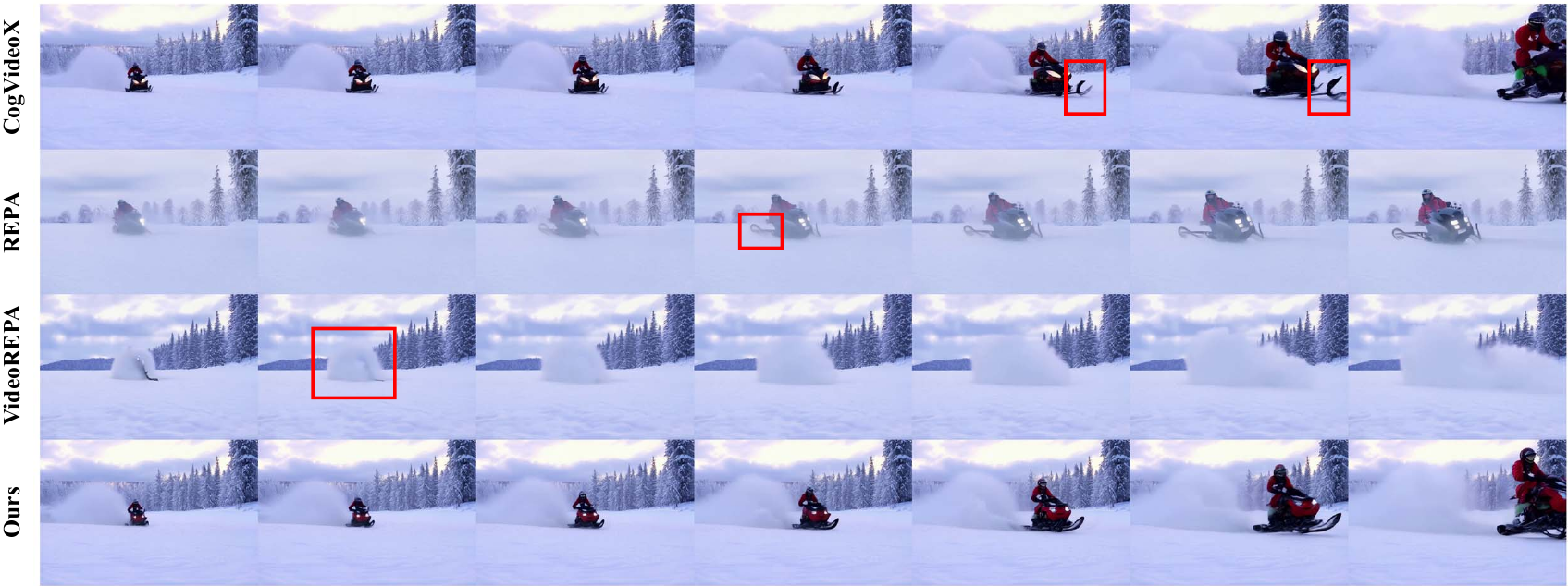}
        \subcaption{Prompt: A snowmobile drives across a snowy lake, kicking up a large spray of snow behind it.}
        \label{fig:supp_qual_2}
    \end{subfigure}

    \vspace{2mm}

    \begin{subfigure}{\linewidth}
        \centering
        \includegraphics[width=\linewidth]{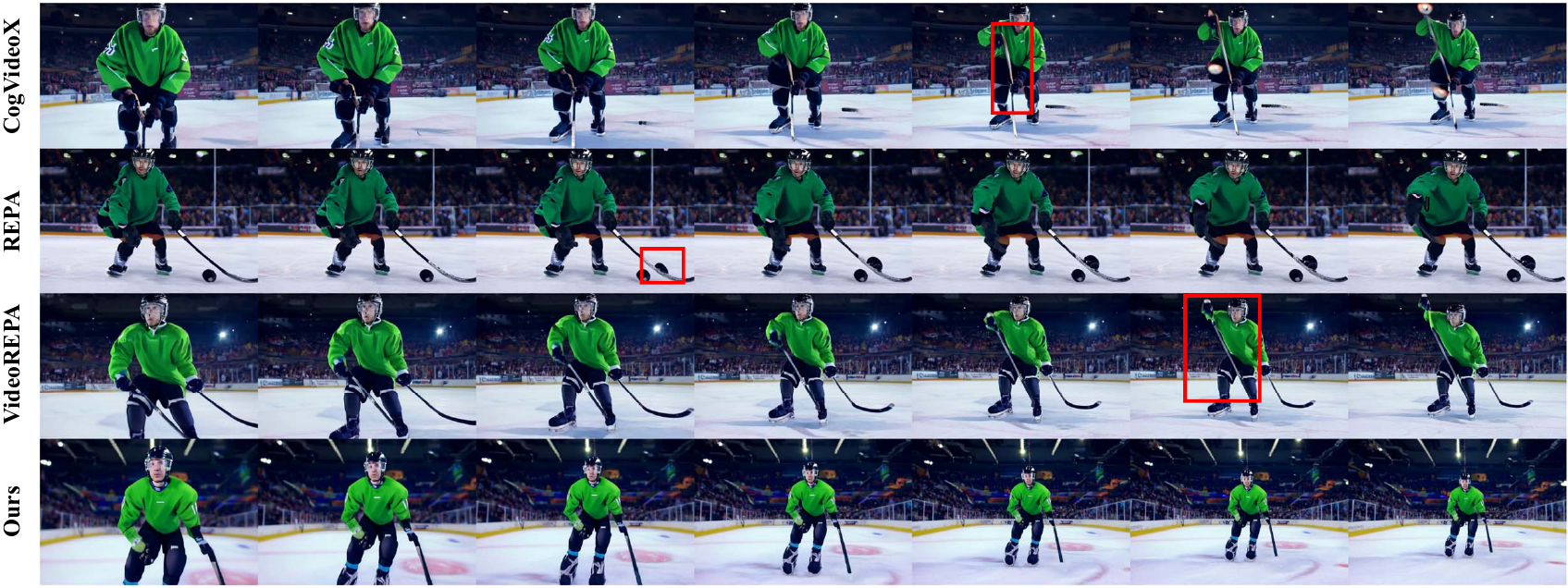}
        \subcaption{Prompt: A player performs a drag flick, hitting the ball forcefully from the ground towards the goal.}
        \label{fig:supp_qual_3}
    \end{subfigure}

    \caption{\textbf{Qualitative results.} Red rectangles highlight regions with physically implausible or temporally inconsistent motion. Our method generates videos with physically realistic dynamics.}
    \label{fig:supp_qual}
\end{figure*}

In Fig.~\ref{fig:supp_qual}, we provide additional qualitative comparisons of videos generated by CogVideoX~\cite{cogvideox}, REPA~\cite{repa}, VideoREPA~\cite{videorepa}, and our method.
In Fig.~\ref{fig:supp_qual_1}, existing methods produce unrealistic dynamics such as floating balls or misplaced shadows, while our method generates physically grounded ball movement.
In Fig.~\ref{fig:supp_qual_2}, competing methods exhibit shape distortion or unnatural snowmobile motion, whereas our method produces temporally consistent and physically plausible motion dynamics.
In Fig.~\ref{fig:supp_qual_3}, existing methods suffer from artifacts such as elongated hockey sticks or spuriously appearing objects, while our method maintains realistic and coherent motion throughout.

Overall, our method consistently improves physical plausibility in generated videos. However, as shown in Fig.~\ref{fig:supp_qual_1} and Fig.~\ref{fig:supp_qual_3}, it still fails to faithfully follow the text prompt.
We attribute this limitation to the relatively small scale of our base model, and expect that applying our method to larger models such as CogVideoX-5B~\cite{cogvideox} or HunyuanVideo~\cite{hunyuanvideo} would alleviate this issue.
We leave this as future work due to computational resource constraints.

\end{document}